# Towards Understanding Language through Perception in Situated Human-Robot Interaction: From Word Grounding to Grammar Induction

**Amir Aly and Tadahiro Taniguchi**

Emergent Systems Laboratory
Ritsumeikan University
Japan
email: {amir.aly@em.ci.ritsumei.ac.jp} & {taniguchi@em.ci.ritsumei.ac.jp}

Robots are widely collaborating with human users in different tasks that require high-level cognitive functions to make them able to discover the surrounding environment. A difficult challenge that we briefly highlight in this short paper is inferring the latent grammatical structure of language, which includes grounding parts of speech (e.g., verbs, nouns, adjectives, and prepositions) through visual perception, and induction of **C**ombinatory **C**ategorial **G**rammar (**CCG**) for phrases. This paves the way towards grounding phrases so as to make a robot able to understand human instructions appropriately during interaction.

Grounding words through visual perception - using a probabilistic generative model - has the objective of making a robot able to understand the meaning of action verbs, object characteristics (i.e., color and geometry), and spatial relationships between objects in space through a cross situational learning context between a human tutor and a robot (without any previous knowledge of language). This implies inducing unsupervised **P**art-**o**f-**S**peech (**POS**) tags representing syntactic categories of words and grounding them, with the meaning of words, through visual perceptual information (Figures 1 & 2).

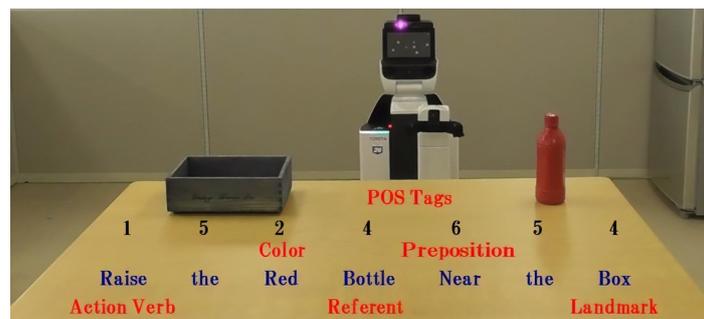

**Fig. (1)** The robot determines the action verb, the referent and its color, and the landmark in the sentence through grounding words and their numerical unsupervised POS tags in visual perception.

However, grounding the induced syntactic categories of words does not illustrate the latent syntactic structure of language and is not sufficient to understand phrases composed of more than one word, which explains the need for investigating a deeper level of syntactic representation of language. To this end, the main contribution of this work aims at inducing **C**ombinatory **C**ategorial **G**rammar (**CCG**) (Steedman 2000) for the phrase and sentence levels based on the grounded syntactic tags of words, which opens the door to understanding phrases (not only words) composing a sentence.

This research topic could help with the challenge of understanding the latent syntactic structure of language that faces the communities of computational linguistics and cognitive robotics. In computational linguistics, the literature reveals different models for grammar induction. However, they were always dependent on annotated databases including syntactic tags of words (e.g., noun, verb, etc.). In other words, these models did not consider grounding words through visual perception as a basis for grammar induction and learning in a developmentally plausible manner (Bisk and Hockenmaier, 2013). Meanwhile, in cognitive robotics, the majority of studies consider the problem of "Symbol Grounding" in language as grounding words only through perception, which does not allow for studying dependencies between words (Tellex et al., 2011) (i.e., they did not investigate grammar understanding at the phrase level).

In this work, we bridge between cognitive robotics and computational linguistics, and propose a framework for grounding lexical information of words through visual perception so as to infer the combinatorial syntactic structure of language - in an unsupervised manner - within a situated human-robot interaction context.

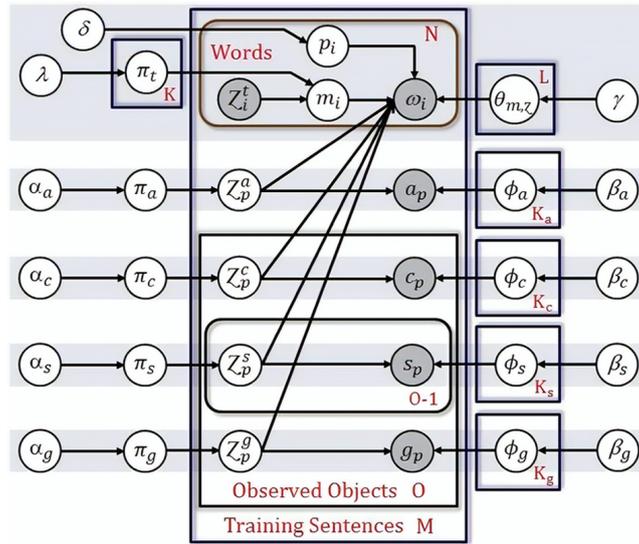

**Fig. (2)** Graphical representation of the grounding model composed of four different modalities of information. The observations $a_p$, $c_p$, $s_p$, $g_p$ denote the characterizing information of **action verbs** (expressed through the tracked coordinates of the human arm joints while doing actions), **object color** (expressed through the color histogram of each segmented object in 3D point cloud), **spatial relationships** between objects (expressed through the coordinates of the segmented objects in 3D point clouds), and **object geometry** (expressed through the VFH descriptor of each segmented object in 3D point cloud). The observation $Z_i$ denotes the induced POS numerical tags of words.

**C**ombinatory **C**ategorial **G**rammar (**CCG**) is an expressive syntactic formalism proposed by Steedman (2000), where any two syntactic categories amongst the atomic (***S***, ***N***, and ***NP***), functor (e.g., ***NP/N***), or modifier (e.g., ***N/N***) categories of neighboring constituents could be combined through a group of rules so as to create complex categories corresponding to higher-level constituents. The slash operators: "/" indicates a forward combination (e.g., an argument follows a functor), and "\" indicates a backward combination (e.g., an argument precedes a functor). The standard unary and binary combinatorial rules of the CCG formalism include:

1. Application combinators:
$$X/Y \quad Y \quad \xRightarrow{\text{Forward } >} \quad X$$

2. Composition combinators:
$$X/Y \quad Y/Z \quad \xRightarrow{\text{Forward } >B} \quad X/Z$$

3. Type-raising combinators:
$$Y \quad \xRightarrow{\text{Forward } >T} \quad X/(X\backslash Y)$$

Figure (3) shows an example to illustrate the use of application combinators to create bottom-up parsing of constituents. Grounding these phrases through perception would allow for understanding the underlying dependencies between words leading to understanding sentences, which is a future scope of this current study.

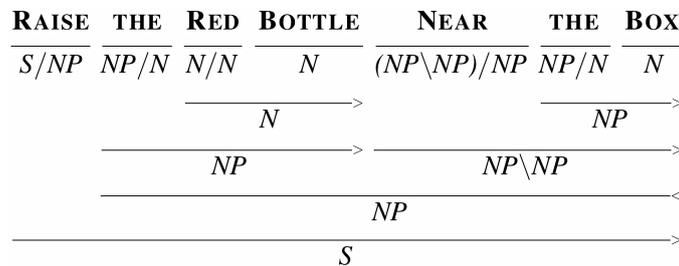

**Fig. (3)** CCG parsing through forward and backward application combinators. The tags ***N*** denotes a noun, ***NP*** denotes a noun phrase, and ***S*** denotes a sentence.

The employed Bayesian nonparametric **HDP-CCG** induction model (Figure 4) uses **D**irichlet **P**rocesses (**DP**) (Teh et al., 2006) to generate an infinite set of CCG categories, defined through stick-breaking processes and multinomial distributions over categories.

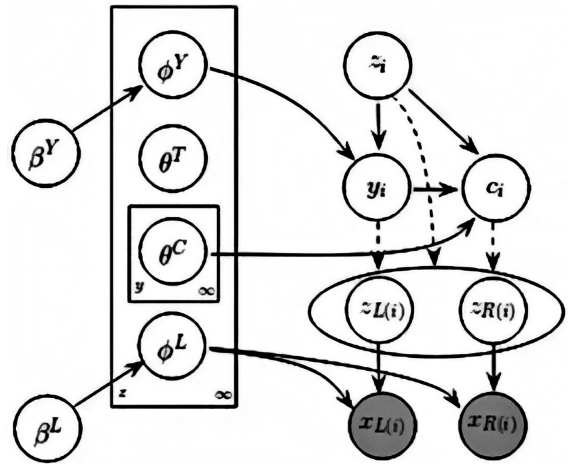

**Fig. (4)** HDP-CCG induction model, where *c* denotes the combinator, *y* denotes the argument, and *z* denotes the parent, which define the left and right children $Z_{Left}$ and $Z_{Right}$.

Briefly, in this research, we propose a framework for grounding syntactic categories of words through visual perception, and inferring the combinatorial syntactic structure of phrases. This could effectively help in grounding phrases and their induced CCG complex categories so as to allow a robot to understand human instructions during interaction, which constitutes a direction of future research. More details on this approach are available in Aly et al., (2017, 2018a,b,c).

**ACKNOWLEDGMENT**

This work was supported by AIP-PRISM, **J**apan **S**cience and **T**echnology Agency "**JST**". Grant number JPMJCR18Z4, Japan.

**References**

[1] M. Steedman, editor, "The Syntactic Process". The MIT Press, Cambridge MA, USA, 2000.
[2] Y. Bisk and J. Hockenmaier, "An HDP Model for Inducing Combinatory Categorial Grammars". Transactions of the Association for Computational Linguistics , 1:75–88, 2013.
[3] S. Tellex, T. Kollar, S. Dickerson, M. R. Walter, A. G. Banerjee, S. Teller, and N. Roy, "Approaching the Symbol Grounding Problem with Probabilistic Graphical Models". AI Magazine , 32 (4):64–76, 2011.
[4] Y-W. Teh, M. I. Jordan, M. J. Beal, and D. M. Blei, "Hierarchical Dirichlet Processes". Journal of the American Statistical Association , 101(476):1566–1581, 2006.
[5] A. Aly, A. Taniguchi, and T. Taniguchi, "A Generative Framework for Multimodal Learning of Spatial Concepts and Object Categories: An Unsupervised Part-of-Speech Tagging and 3D Visual Perception based Approach". In Proceedings of the 7th Joint IEEE International Conference on Development and Learning and on Epigenetic Robotics (ICDL-EpiRob) , Lisbon, Portugal, 2017.
[6] A. Aly and T. Taniguchi, "Towards Understanding Object-Directed Actions: A Generative Model for Grounding Syntactic Categories of Speech through Visual Perception". In Proceedings of the IEEE International Conference on Robotics and Automation (ICRA), Brisbane, Australia, 2018a.
[7] A. Aly, T. Taniguchi, and D. Mochihashi, "A probabilistic Approach to Unsupervised Induction of Combinatory Categorial Grammar in Situated Human-Robot Interaction". In Proceedings of the 18th IEEE-RAS International Conference on Humanoid Robots (Humanoids), Beijing, China, 2018b.
[8] O. Roesler, A. Aly, T. Taniguchi, and Y. Hayashi, "A Probabilistic Framework for Comparing Syntactic and Semantic Grounding of Synonyms through Cross-situational Learning". In Proceedings of the International Workshop on Representing a Complex World: Perception, Inference, and Learning for Joint Semantic, Geometric, and Physical Understanding, in Conjunction with the IEEE International Conference on Robotics and Automation (ICRA), Brisbane, Australia, 2018c.
[9] O. Roesler, A. Aly, T. Taniguchi, and Y. Hayashi, "Evaluation of Word Representations in Grounding Natural Language Instructions through Computational Human-Robot Interaction", Proceedings of the 14th ACM/IEEE Human-Robot Interaction Conference (HRI), South Korea, 2019.
[10] A. Aly, T. Taniguchi, and D. Mochihashi, "A Bayesian Approach to Phrase Understanding through Cross-Situational Learning", Proceedings of the International Workshop on *Visually Grounded Interaction and Language (ViGIL)*", in Conjunction with the 32nd Conference on Neural Information Processing Systems (NeurIPS), Canada, 2018.
[11] A. Aly, T. Taniguchi, and D. Mochihashi, "Towards Understanding Syntactic Structure of Language in Human-Robot Interaction", Proceedings of the International Workshop on *Visually Grounded Interaction and Language (ViGIL)*", in Conjunction with the 32nd Conference on Neural Information Processing Systems (NeurIPS), Canada, 2018.